\pdfoutput=1

\documentclass[11pt]{article}[final]

\usepackage[dvipsnames]{xcolor}

\usepackage{acl}
\usepackage{nert}

\usepackage{pgfplots}
\pgfplotsset{compat=1.17}

\usepackage{times}
\usepackage{latexsym}
\usepackage{graphicx}
\graphicspath{ {./images/} }
\usepackage{linguex}
\usepackage{tabularx}

\usepackage[utf8]{inputenc}

\usepackage{bbm}
\usepackage{subcaption}

\renewcommand{\mathbbm}[1]{\text{\usefont{U}{bbm}{m}{n}#1}} 

\newcommand{\words}{\textcolor{\mathcolour}{\ensuremath{\mathbf{w}}}\xspace}
\newcommand{\word}{\textcolor{\mathcolour}{\ensuremath{w}}\xspace}
\newcommand{\context}{\textcolor{\mathcolour}{\ensuremath{\mathbf{c}}}\xspace}
\newcommand{\nwords}{\textcolor{\mathcolour}{\ensuremath{N}}\xspace}
\newcommand{\params}{\textcolor{\mathcolour}{\ensuremath{\theta}}\xspace}
\newcommand{\unigram}{\textcolor{\mathcolour}{\ensuremath{U}}\xspace}
\newcommand{\alphabet}{\textcolor{\mathcolour}{\ensuremath{\Sigma}}\xspace}
\newcommand{\slor}{\textcolor{\mathcolour}{\ensuremath{\mathcal{S}}}\xspace}
\newcommand{\deltaslor}{\textcolor{\mathcolour}{\ensuremath{\Delta\mathcal{S}}}\xspace}

\newcommand{\condname}{Manip}
\newcommand{\ltaln}{LtAln}

\newcommand{\gramsent}{\textcolor{\mathcolour}{\ensuremath{\suiteitem \, _{\config{-}{\condname}}^{\ell}}}\xspace}
\newcommand{\ungramsent}{\textcolor{\mathcolour}{\ensuremath{\suiteitem \, _{\config{+}{\condname}}^{\ell}}}\xspace}
\newcommand{\suiteitem}{\textcolor{\mathcolour}{\ensuremath{i}}\xspace}

\newcommand{\config}[2]{\ensuremath{#1\textsc{#2}}\xspace}

\newcommand{\letalone}{\textcolor{black}{\textsc{let-alone}}\xspace}

\renewcommand{\nertcomment}[4]{\unskip}

\title{Unpacking \textit{Let Alone}: Human-Scale Models Generalize to a Rare Construction in Form but not Meaning}

\author{Wesley Scivetti \quad Tatsuya Aoyama \quad Ethan Wilcox \quad Nathan Schneider \\
  Georgetown University \\
  \{\emldisplay{wss37@georgetown.edu}{wss37},
  \emldisplay{ta571@georgetown.edu}{ta571},
  \emldisplay{ethan.wilcox@georgetown.edu}{ethan.wilcox},
  \emldisplay{nathan.schneider@georgetown.edu}{nathan.schneider}\}\texttt{@georgetown.edu} \\
}
\date{}

\begin{document}
\maketitle
\begin{abstract}

Humans have a remarkable ability to acquire and understand grammatical phenomena that are seen rarely, if ever, during childhood. 
Recent evidence suggests that language models with human-scale pretraining data may possess a similar ability by generalizing from frequent to rare constructions.
However, it remains an open question how widespread this generalization ability is, and to what extent this knowledge extends to \emph{meanings} of rare constructions, as opposed to just their \emph{forms}. 
We fill this gap by testing human-scale transformer language models on their knowledge of both the form and meaning of the (rare and quirky) English \letalone construction.
To evaluate our LMs we construct a bespoke synthetic benchmark that targets syntactic and semantic properties of the construction.
We find that human-scale LMs are sensitive to form, even when related constructions are filtered from the dataset.
However, human-scale LMs do not make correct generalizations about \letalone's meaning.
These results point to an asymmetry in the current architectures' sample efficiency between language form and meaning, something which is not present in human language learners.\footnote{Code and data: \url{https://github.com/WesScivetti/BabyAlone}}

\end{abstract}

\section{Introduction}

The ability of neural network--based language models (LMs) to learn human language has profound implications for our theories of learning and cognitive science of language \citep{warstadt2022artificial, wilcoxUsingComputationalModels2024}.
Of particular interest is whether LMs trained on human-scale data can learn nuanced humanlike generalizations about linguistic form and meaning \citep{wilcox2025bigger}.
Recent studies have found that models have remarkable success at both, but that human-scale models appear to make better generalizations about linguistic form \citep{warstadtFindingsBabyLMChallenge2023}.
This is particularly true when it comes to rare constructions, where models have been shown to learn formal constraints much more robustly than constraints about the constructions’ meanings \citep{Weissweiler_Hofmann_Köksal_Schütze_2022}.

These results pose a potential problem for construction-based theories of grammar.
Construction grammar is a family of theories, which propose that linguistic knowledge is stored in templatic packets (constructions), and that a constructions’ form and meaning are learned simultaneously \citep{Goldberg_2006}.
These theories predict that if constructionist learning is happening in LMs, form and meaning should be learned simultaneously.
However, previous studies that test form and meaning of specific constructions in a controlled way \citep[e.g.,][]{Weissweiler_Hofmann_Köksal_Schütze_2022} do so only in large-scale LMs, which limits their cognitive interpretation \citep{wilcox2025bigger}.
Other studies, which assess human-scale LMs, investigate formal and semantic competence on different phenomena \citep{warstadtFindingsBabyLMChallenge2023}.

In this work, therefore, we test for a form--meaning learning asymmetry on human-scale LMs.
We focus on the \letalone construction (\cref{sec:let-alone}), a rare construction of English that is subject to a nuanced but well-studied set of syntactic, semantic and pragmatic constraints \citep{fillmoreRegularityIdiomaticityGrammatical1988}.
We train human-scale models from scratch and assess them using a bespoke behavioral test suite that probes various facets of \letalone, including conjunction, negation, and scalar properties (\cref{sec:methods}).
Our experiments show strong evidence for a form--meaning asymmetry: Human-scale models learn \letalone's formal constraints almost perfectly (\cref{sec:exp1-formal}), but fail to learn any semantic constraints (\cref{sec:exp2-semantic}).
Moreover, we find that our models still learn \letalone{}’s formal constraints \emph{even when examples of \textsc{let-alone} and related constructions are filtered from pretraining data} (\cref{sec:exp3-filtered}). These results underscore the claim that indirect evidence can be crucial for learning constraints on rare constructions at human-scale \citep{misraLanguageModelsLearn2024}. However, we find that removing instances of individual \textit{let} and \textit{alone} tokens from pretraining altogether destroys sensitivity to most formal constraints. An emerging theme in the literature on LM learning of constructions (\cref{sec:related}), the performance disparity between form and meaning calls into question how much human-scale models can learn about the meaning of rare constructions.

\begin{table*}[]
    \centering\smaller
    \def\arraystretch{1.25}
    \setlength{\tabcolsep}{4pt}
    \begin{tabular}{@{}lp{.15\textwidth}rp{.22\textwidth}p{.42\textwidth}@{}}
        \toprule
        \textbf{Test Type} & \textbf{Property} & \multicolumn{1}{c}{\textbf{N}} & \textbf{Manipulation} & \textbf{Example}\\
        \midrule
        \textsc{Formal} & Conjunction (Clause) & \hphantom{0}5217 & \textit{conjoin} independent clauses & *I couldn't lift the blue crate let alone I couldn't lift the red crate.\\
        \textsc{Formal} & Conjunction (VP) & \hphantom{0}5217 & \textit{conjoin} VPs & I couldn't lift the blue crate let alone lift the red crate.\\
        \textsc{Formal} & Conjunction (Gap) & \hphantom{0}5217 & \textit{conjoin} elided VP clauses & I couldn't lift the blue crate let alone you the red crate.\\
        \textsc{Formal} & Clefting & \hphantom{0}5217 & \textit{cleft} S & *It is the blue crate let alone the red crate that I couldn't lift.\\
        \textsc{Formal} & NPI & \hphantom{0}5217 & \textit{remove} ``not'' & *I could lift the blue crate let alone the red crate.\\
        
        \textsc{Meaning} & Scalar Semantics & 16887 & \textit{contradictory follow-up} & \#I couldn't lift the blue crate let alone the red crate. The blue crate is heavier than the red crate. \\
        \bottomrule
    \end{tabular}
    \caption{Number of examples (N) for each test in our test set for \textsc{let-alone}. These tests are inspired by the properties of \textsc{let-alone} as described in \citet{fillmoreRegularityIdiomaticityGrammatical1988}. An example for each manipulation is shown relative to the base sentence \pex{I couldn't lift the blue crate let alone the red crate.} Some of the manipulations induce an ungrammatical \letalone sentence where \textit{and} would be acceptable, while others are equally acceptable in the base and manipulated versions.}
    \label{tab:dataset_stats}
\end{table*}

\section{Background: \emph{Let Alone}}\label{sec:let-alone}
In this work, we focus our attention on the \letalone construction. This construction was analyzed in detail by \citet{fillmoreRegularityIdiomaticityGrammatical1988}. The \textsc{let-alone} construction joins two constituents, which can be of various types, as shown in \ref{ex:one}--\ref{ex:three}:

\ex.\label{ex:one} Max won't eat shrimp, let alone squid. \textsc{[NP Conjunction]}

\ex.\label{ex:two} I barely got up in time to cook lunch, let alone cook breakfast. \textsc{[VP conjunction]}

\ex.\label{ex:three} They couldn't make John eat the shrimp, let alone Lucille the squid. \textsc{[Elided VP conjunction]}




However, unlike the prototypical conjoiner, \emph{and}, \textsc{let-alone} cannot join two full, unelided independent clauses, as in \cref{ex:conj1}.


\ex.\label{ex:conj1} *I couldn't afford the red sunglasses let alone I couldn't afford the black sunglasses.



Additionally, \textsc{let-alone} resists movement and fronting in situations which generally allow it. For example, \textsc{let-alone} can't be inserted into cleft sentences, like in \cref{ex:cleft1}.

\ex.\label{ex:cleft1} *It is the red sunglasses let alone the black sunglasses that I couldn't afford.

\textsc{let-alone} is also considered a negative polarity item (NPI), and generally ungrammatical when not under the scope of negation, as in \cref{ex:npi1}.\footnote{Though see \citet{fillmoreRegularityIdiomaticityGrammatical1988} for some examples where NPI licensors are not needed, such as ``You have enough material for a semester, let alone a week.''} 

\ex.\label{ex:npi1} *I could lift the orange crate let alone the green crate.


Regarding the semantics of \textsc{let-alone}, \citet{fillmoreRegularityIdiomaticityGrammatical1988} argue that it is best understood as one member of a family of paired focus constructions, including ``never mind'', ``much less'', and ``not to mention''. These constructions have two phrases that are simultaneously placed in focus and are semantically connected via a comparison. Specifically, \letalone implies that the two phrases in focus are in a scalar relationship. The two phrases are thus interpreted as being ``two points on a scale'' \citep{fillmoreRegularityIdiomaticityGrammatical1988}, with the second phrase being higher than the first phrase on whatever scale is evoked. In this way, \textsc{let-alone} has some semantic shared properties with more general comparative constructions, which also place two entities on either explicit or implied scales. 

Importantly for our studies, \letalone is exceedingly rare. In our pretraining corpus of $\approx$100 million words, \letalone occurs fewer than 400 times. To our knowledge, \textsc{let-alone} is the most infrequent construction to be examined in a study such as this at human-scale.

In the experiments that follow, we test both syntactic and semantic properties of \letalone. 
While descriptions of our test items are given in the respective experimental sections, at a high level, we test four properties:
For formal properties, we test NPI licensor sensitivity, clefting, and restrictions on conjunction. For semantic properties, we test scalar sensitivity, specifically, whether models prefer contexts whose properties match the scales of \letalone.


\section{Methods}\label{sec:methods}

\newcommand{\mathcolour}{BrickRed}
\newcommand{\nitems}{\textcolor{\mathcolour}{\ensuremath{k}}\xspace}

\subsection{Evaluation Dataset}

We experiment on a templatically constructed dataset of \textsc{let-alone} minimal pair instances. This approach follows other minimal pair datasets which are used to test LM processing of grammatical phenomena (e.g.~SyntaxGym, \citealp{gauthierSyntaxGymOnlinePlatform2020b}; BLiMP, \citealp{warstadtBLiMPBenchmarkLinguistic2020b}; COMPS, \citealp{Misra_Rayz_Ettinger_2023}). 
For each of our experiments, we create a test suite of \nitems templatically generated items.
Each item comes in four different conditions, which cross a \emph{grammatical manipulation} (\config{\pm}{\condname}) with the \emph{presence or absence} of \letalone (\config{\pm}{\ltaln}). 
\Cref{ex:conditions_example} serves as an example:

\ex. \label{ex:conditions_example}
\a. * Max \textbf{could} lift the red box \textbf{let alone} the blue box. [\config{+}{{\condname}}, \config{+}{\ltaln}\unskip] \label{ex:a}
\b. Max \textbf{couldn't} lift the red box \textbf{let alone} the blue box. [\config{-}{\condname}, \config{+}{\ltaln}\unskip] \label{ex:b}
\c. Max \textbf{could} lift the red box \textbf{and} the blue box. [\config{+}{{\condname}}, \config{-}{\ltaln}\unskip] \label{ex:c}
\d. Max \textbf{couldn't} lift the red box \textbf{and} the blue box. [\config{-}{{\condname}}, \config{-}{\ltaln}\unskip] \label{ex:d}

Note that \config{+}{\condname} is a grammatical manipulation that makes \letalone sentences (\config{+}{\ltaln}) \textit{ungrammatical} (for some manipulations), but does not affect the grammatically of non-\letalone sentences (\config{-}{\ltaln}); hence, only the [\config{+}{Manip}, \config{+}{\ltaln}\unskip] configuration is ever ungrammatical. The specific manipulations we test involve conjoining independent clauses, clefting the sentence, and removing the negative licensor (``not''), respectively, as shown in \Cref{tab:dataset_stats}. We also experiment with two additional conjunction experiments for which the manipulations of \textsc{let-alone} are grammatical: conjoining verb phrases (VPs) and conjoining elided VPs.
From the sentences in these five conditions, we calculate an \emph{accuracy score}, which is described in the next section in detail. We opt for template-based examples, as opposed to natural corpus data, in order to control for several factors, including the frequency of lexical items, the scalar semantics invoked by \textsc{let-alone}, and the syntactic context in which the \textsc{let-alone} sentence occurs. 

There is evidence that human-scale models struggle with world knowledge \citep{ivanovaElementsWorldKnowledge2024,huFindingsSecondBabyLM2024}, and thus their performance on interpreting the scalar semantics of \textsc{let-alone} in examples like \cref{ex:one,ex:two,ex:three} may be bottlenecked by their lack of reasoning over the properties which are being compared on the scale (e.g., the unusualness of eating shrimp versus squid). Because of this, we design templates around scalar properties from domains which involve quantitative scales, such as height, weight, distance, and price. For consistency, all of our \letalone focus elements are direct objects. To control for the possibility of lexical biases inherent to the objects, our focus elements are always the same lexical noun (e.g. ``box'' in the example above), which is modified with different neutral adjectives, such as color terms. \cref{tab:dataset_stats} reports example counts for the dataset.

The meaning of \textsc{let-alone} implies that the two phrases in the construction have some shared scalar property, and the second one has a higher value than the first. This allows us to probe the semantics of the construction by designing minimal pairs in which there is a follow-up sentence which either directly follows from the scalar semantics of the construction, or contradicts it, while still maintaining our overall 2x2 manipulation: 

\ex. I couldn't afford the red sunglasses \textbf{let alone} the black sunglasses. \label{ex:sem1}
\a. \# The \textbf{red} sunglasses are more expensive than the \textbf{black} sunglasses. [\config{+}{{\condname}}, \config{+}{\ltaln}\unskip]
\b. The \textbf{black} sunglasses are more expensive than the \textbf{red} sunglasses. [\config{-}{{\condname}}, \config{+}{\ltaln}\unskip]

\ex. I couldn't afford the red sunglasses \textbf{and} the black sunglasses. \label{ex:sem2}
\a. The \textbf{black} sunglasses are more expensive than the \textbf{red} sunglasses. [\config{+}{{\condname}}, \config{-}{\ltaln}\unskip]
\b. The \textbf{red} sunglasses are more expensive than the \textbf{black} sunglasses. [\config{-}{{\condname}}, \config{-}{\ltaln}\unskip]



The only difference in the manipulated examples is that the color items have been swapped. This shouldn't impact models' predictions when the context sentence contains \emph{and}.  However, the \textsc{let-alone} sentence makes clear which object is higher on the expense scale, and so only one follow-up sentence is valid, while the other is infelicitous. 


\subsection{Evaluation Metric}

Following \citet{misraLanguageModelsLearn2024}, instead of comparing raw probabilities between conditions, we use the Syntactic Log Odds Ratio \citep[SLOR;][]{Pauls_Klein_2012, lau2017grammaticality}.
We calculate SLOR over sentences, $\words=[\word_1 \dots \word_{\nwords}]$, where \word is drawn from a vocabulary of words \alphabet. A sentence $\words$ is potentially conditioned on a context $\context$. In our form experiments, $\context$ is empty, and $\words$ is the \letalone sentence. In our meaning experiment, the context $\context$ is our \letalone sentence, and $\words$ is a following sentence that is either felicitous or non-felicitous given the scalar properties of the \letalone context.
We assume an LM with parameters \params that can produce probability $p_{\params}(\cdot)$, and a model $p_{\unigram}(\cdot)$ of the unigram distribution over $\word \in \alphabet$.
SLOR can then be defined as:
\begin{align}
    \slor(\words) = \frac{1}{\nwords} \log \frac{p_{\params}(\words \mid \context)}{\prod_{\word \in \words} p_{\unigram}(\word)}
\end{align}

SLOR is designed to control for the fact that more frequent words are inherently less surprising for an LM. 
To turn by-sentence SLOR scores into accuracy scores, we compare conditions in our test suites.
Every item in our test suites come in four conditions, corresponding to the examples in \Cref{ex:a}--\Cref{ex:d}.
For a given item, \suiteitem, we notate its conditions with sub and superscripts: $\suiteitem \, _{\config{+}{\condname}}^{\config{+}{\textsc{\ltaln}}}$ refers to \suiteitem's  \config{+}{{\condname}}, \config{+}{\ltaln} condition.
First, we define what we refer to as the delta SLOR, or \deltaslor, which is simply the difference in SLOR due to our grammatical manipulation: 
\begin{align}
    \deltaslor_{\suiteitem}&(\ell) = \slor(\gramsent) - \slor(\ungramsent)
\end{align}

\noindent where $\ell$ can either be $+$\textsc{\ltaln} or $-$\textsc{\ltaln}.
To calculate accuracy, we inspect the \emph{differences} in \deltaslor scores.
Namely, we predict that the effect of grammatical manipulation should be greater (reflecting lower grammaticality) when \letalone is used, compared to when a vanilla conjunction is used.
With this prediction, our accuracy scores for a test suite of \nitems items can be defined as:
\begin{align}
    \frac{1}{\nitems} \sum_{i=1}^{\nitems} \mathbbm{1}[\deltaslor_{\suiteitem}(\textsc{$+$\ltaln}) \geq \deltaslor_{\suiteitem}(\textsc{$-$\ltaln})]
\end{align}

\noindent This corresponds to the effect of the manipulation in the \letalone case (between \Cref{ex:a} vs.~\Cref{ex:b}) above and beyond that observed with the non-\letalone control (\Cref{ex:c} vs.~\Cref{ex:d}). 

Intuitively, a model achieving high accuracy on this task has correctly understood that these manipulations are more ungrammatical for \textsc{let-alone} than they are for \textit{and}, and has thus learned a core part of the idiosyncratic nature of the construction. We also control for the possible bias related to ordering of colors by swapping the orders of the colors for each example, and an example is only considered correct if \textit{both} orders are correct. Since this involves two pairwise comparisons (both orderings), chance performance on our tasks is 25\%.

\begin{figure*}[ht] 
  \begin{subfigure}[b]{0.5\textwidth}
    \centering 
    \textbf{(a)}
    \includegraphics[width=0.95\textwidth]{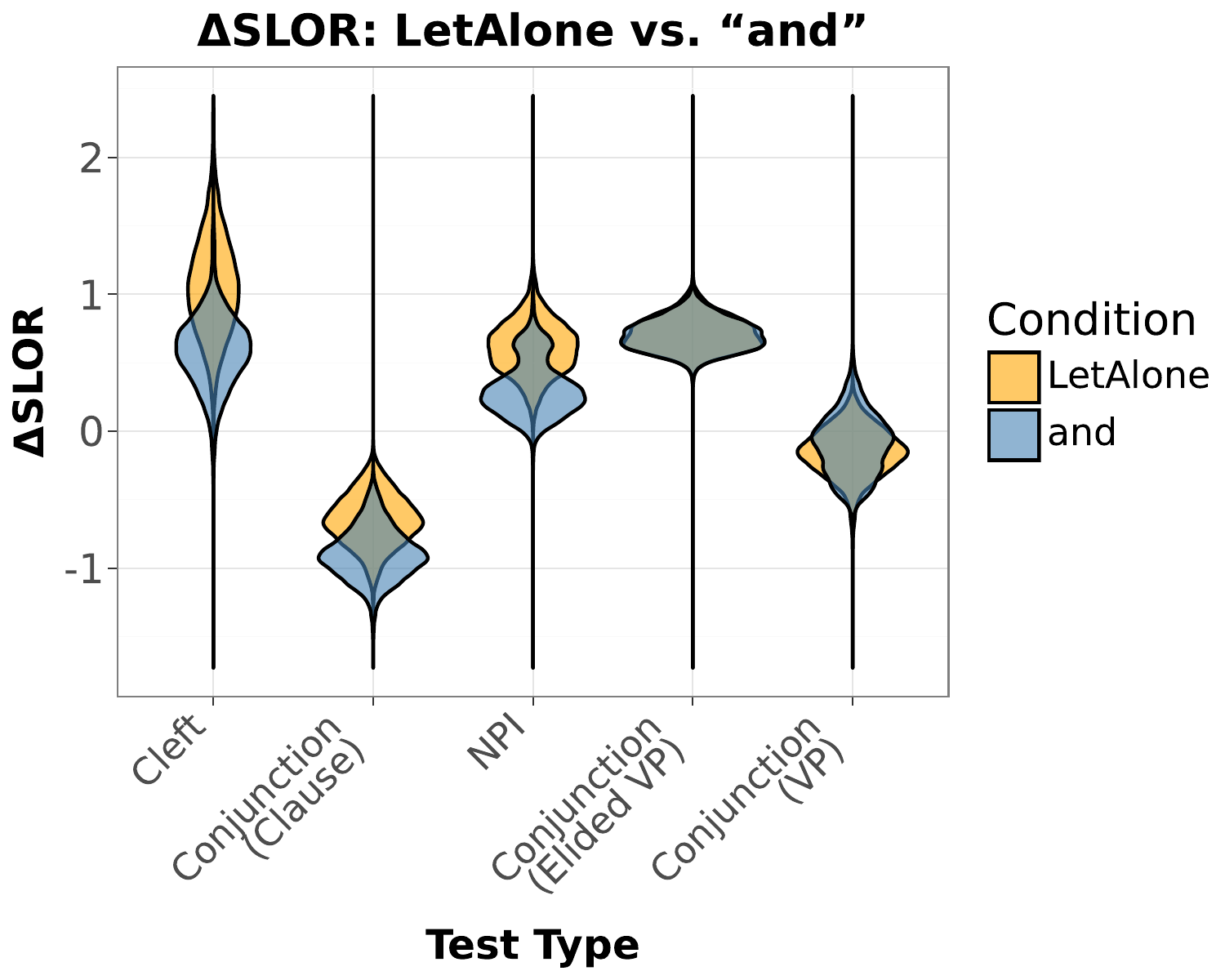}
        \label{fig:violin_slor}
  \end{subfigure}
  \begin{subfigure}[b]{0.5\textwidth}
    \centering 
    \textbf{(b)}
    \includegraphics[width=0.95\textwidth]{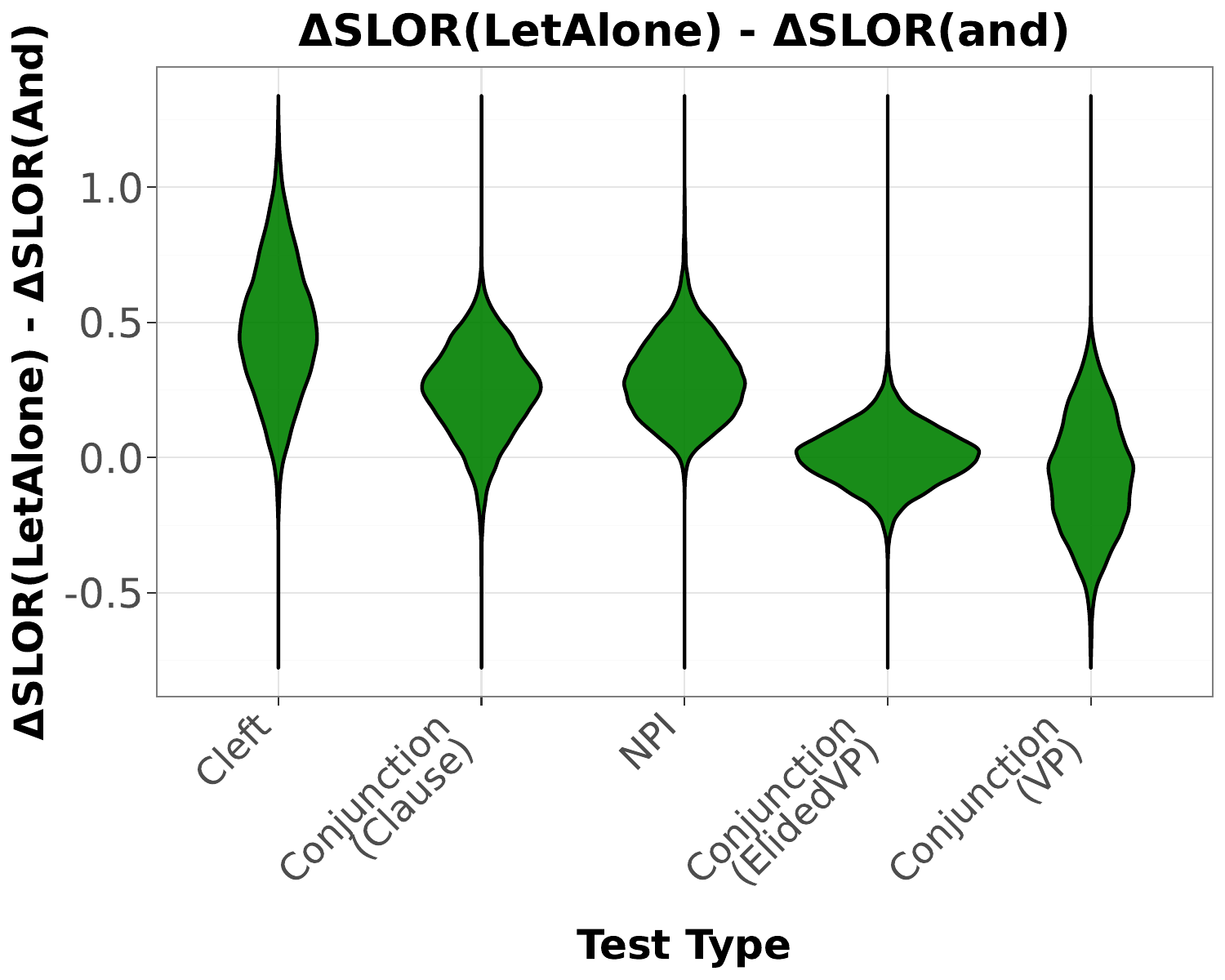}
        \label{fig:deltdelt_slor}
  \end{subfigure}
  \vspace{-0.5cm}
  \caption{\textbf{Results on Syntactic Tests:} \textbf{(a)} shows \deltaslor where higher delta values indicate a greater effect of the constraint. Patterns are consistent with the grammaticality of the syntactic manipulation. \textbf{(b)} shows \deltaslor(LetAlone)- \deltaslor(And).}
  \label{fig:violin_slor2}
\end{figure*}

\subsection{Model Architecture and Pretraining}

We use the training split of the BabyLM-strict 100M dataset \citep{warstadtFindingsBabyLMChallenge2023} for pretraining. For all experiments, we follow \citet{misraLanguageModelsLearn2024} in utilizing the OPT architecture \citep{Zhang_Roller_Goyal_Artetxe_Chen_Chen_Dewan_Diab_Li_Lin_etal._2022}. We utilize identical hyperparameters to those reported in \citet{misraLanguageModelsLearn2024} where possible.\footnote{Full hyperparameters are reported in \cref{tab:model_architecture} in \Cref{hyper}.} For all model settings, we pretrain two models with identical hyperparameters and different random seed, and report the average results.

In all experiments, we use the minicons library \citep{misra2022minicons} for calculating sequence-level surprisals. Following \citet{misraLanguageModelsLearn2024}, we calculate SLOR using the surprisal values from minicons combined with the unigram frequencies from the BabyLM training set.\footnote{In filtered pretraining experiments, we calculate unigram frequencies on the filtered training dataset.}

\section{Experiment 1: Formal Constraints}\label{sec:exp1-formal}

We first test constraints on the formal characteristics of \textsc{let-alone}. We focus on restrictions of \letalone which differ from those for simple conjunctions, namely conjunction of clauses, clefting, and NPI licensor sensitivity. Additionally, we add two conjunction conditions (conjunction of VPs and conjunction of elided VP clauses), which are grammatical for both \textsc{let-alone} and simple conjunctions. These two additional conditions serve as a control to see if the models not only recognize constraints on \textsc{let-alone}, but also recognize valid syntactic variations.










\begin{table}[]
    \centering\small
    \begin{tabular}{ccc}
        \toprule
        \textbf{Formal Property} & \textbf{Prediction} & \textbf{Accuracy} \\
        \midrule
        Conjunction (Clause) & near 100\% & 88.1 $\pm$ \hphantom{0}.8\% \\
        Clefting & near 100\% & 96.5 $\pm$ \hphantom{0}.5\% \\
        NPI & near 100\%  & 98.6 $\pm$ \hphantom{0}.3\%  \\
        \midrule
        Conjunction (VP) & near 25\% & 31.1 $\pm$ 1.3\% \\
        Conjunction (Gap) & near 25\% & 37.5 $\pm$ 1.3\% \\
        \bottomrule
    \end{tabular}
    \caption{ \textbf{Results for Syntactic Tests}: \emph{Prediction} column indicates expected accuracies if the model had made the humanlike generalization. Model accuracies are means over two random seeds. We report 95\% confidence intervals over the means of the two runs. 
    }
    \label{tab:syntax_res}
\end{table}

\subsection{Results}\label{sec:form_results}



Results for these tests are reported in \Cref{tab:syntax_res} and visualized in \cref{fig:violin_slor2}.
Across all 3 formal constraint tests involving ungrammatical manipulations, accuracy is very high over two randomly seeded pretraining runs. For control conditions (VP conjunction and elided VP clause conjunction), accuracy is near chance, as expected. Looking at SLOR differences at the example level in \cref{fig:violin_slor2}, we see strong separation in the $\deltaslor$ scores for \textsc{let-alone} versus \textit{and}. We can see in \cref{fig:violin_slor2} that the few negative $\deltaslor$ difference values for non-control conditions tend to be clustered very close to 0. For the control conditions, $\deltaslor$ differences are generally evenly distributed above and below 0. These results provide evidence that human-scale models are able to capture the formal properties of \textsc{let-alone} well. This strong performance is despite the fact that the \textsc{let-alone} construction only occurs roughly 300 times in the BabyLM training corpus. With so few training examples, it seems likely that for \textsc{let-alone}, indirect evidence is far more important than direct evidence for the recognition of these formal constraints. See \cref{sec:exp3-filtered} for more discussion of filtered pretraining of \textsc{let-alone} and related constructions.  

\section{Experiment 2: Semantic Constraints}\label{sec:exp2-semantic}

Having shown that human-scale models have sensitivity to a range of formal constraints on the \textsc{let-alone} construction, we evaluate whether BabyLM scale language models are sensitive to the scalar semantics of \textsc{let-alone}. As stated in \cref{sec:let-alone}, we test the scalar semantics by supplying additional follow-up sentences, which are either felicitous or non-felicitous with the scale set up by the \letalone construction. We then compare the SLOR values of the two target sentences, conditioned on the \letalone prefixes.




\subsection{Results}\label{sec:sem_results}

\begin{figure}[ht] 
  \begin{subfigure}[b]{0.45\columnwidth}
    \centering 
    \textbf{(a)}
    \includegraphics[width=0.99\textwidth]{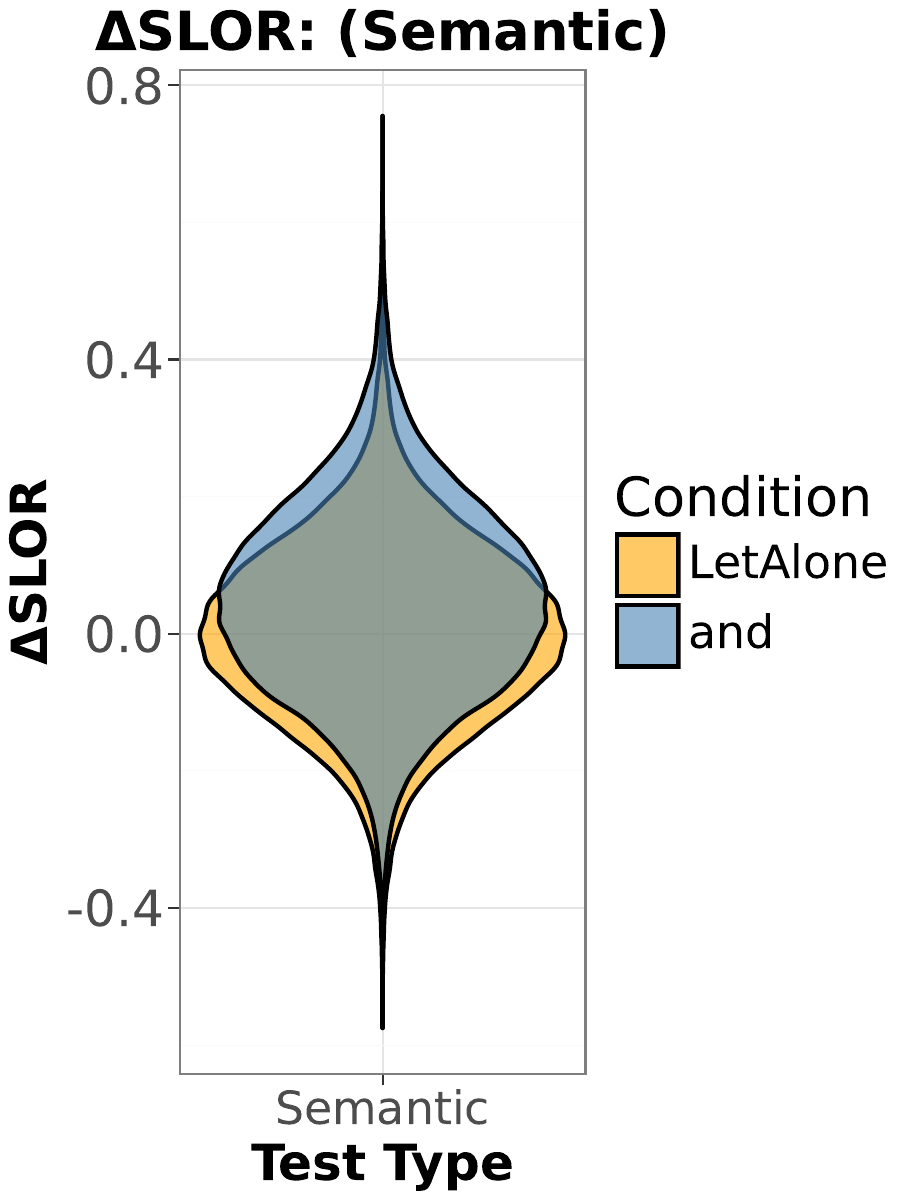}
        \label{fig:violin_slor_sem}
  \end{subfigure}
  \begin{subfigure}[b]{0.45\columnwidth}
    \centering 
    \textbf{(b)}
    \includegraphics[width=0.99\textwidth]{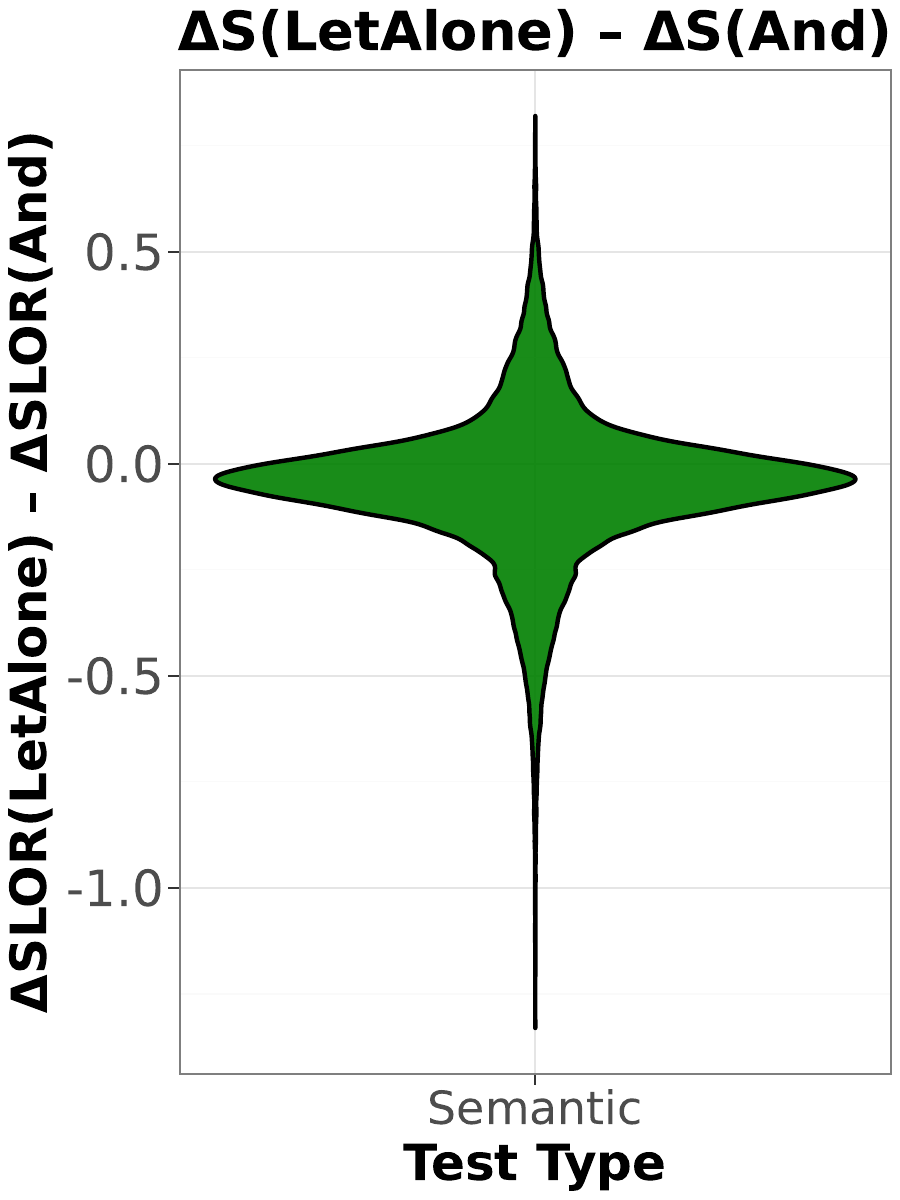}
        \label{fig:deltdelt_slor_sem}
  \end{subfigure}
  \vspace{-0.2cm}
  \caption{\textbf{Results on Semantic Tests:} \textbf{(a)} shows \deltaslor where higher delta values indicate a greater effect of the constraint. Patterns are consistent with the grammaticality of the syntactic manipulation. \textbf{(b)} shows \deltaslor(LetAlone)- \deltaslor(And).}
\end{figure}




\cref{tab:sem_res} reports results on the semantic tests. We find no evidence that BabyLMs are sensitive to the semantics of \textsc{let-alone}, as performance on this minimal pair task is below chance for both random seeds. In contrast, we include as a skyline comparison point GPT-4.1 \citep{openai2024gpt41}, which achieves extremely high accuracy (94\%) on a prompting version of our task on the same dataset.\footnote{See \Cref{gpt_exp} for details on the GPT prompting experiment.} In \cref{fig:violin_slor_sem}, we visualize the $\deltaslor$ values between the correct and incorrect followups, showing that they cluster very close to zero. This indicates that the BabyLM model has very little preference between the two alternatives, pointing to a general lack of sensitivity to the scalar properties of the construction.

\begin{table}[]
    \centering\smaller
    \begin{tabular}{cccc}
        \toprule
        \textbf{Model} & \textbf{Property} & \textbf{Prediction} & \textbf{Accuracy} \\
        \midrule
        BabyLM & Scalar & near 100\% & 4.9 $\pm$ 0.32\% \\
        & Semantics & & \\
        \midrule
        GPT-4.1 & Scalar & near 100\% & 94.0 $\pm$ 0.361\%  \\
        & Semantics & & \\
        \bottomrule
    \end{tabular}
    \caption{\textbf{Accuracies for the Semantic Tests.} BabyLM models demonstrate no sensitivity to the scalar semantics of \letalone. In a metalinguistic prompting formulation of our task, GPT-4.1 achieves strong performance, indicating the dataset is solvable with sufficient input.}
    \label{tab:sem_res}
\end{table}

\subsection{Analysis}

\begin{figure}
    \centering
    \includegraphics[width=\columnwidth]{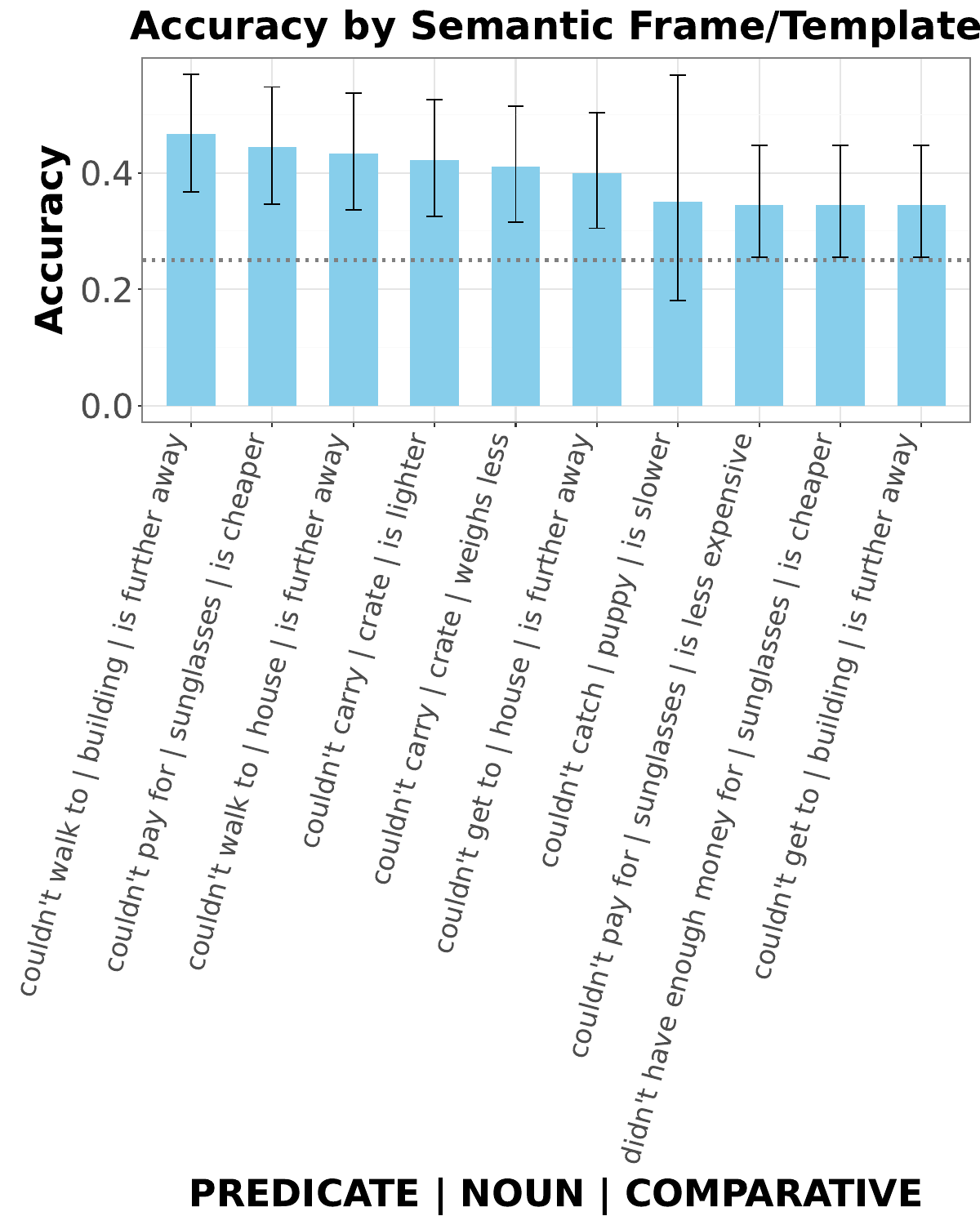}
    \caption{\textbf{Top 10 Accuracies on the Semantic Tests} when separated by predicate, noun, and comparative that fill the template. Error bars indicate 95\% confidence intervals over the results of two random seeds. Above-chance accuracies indicate that the model has some nontrivial semantic performance on that template.}
    \label{fig:acc_by_item}
\end{figure}

We perform an analysis to see what contributes to the poor semantic performance for BabyLM models. We hypothesize that the specific template we use may impact our results. In \cref{fig:acc_by_item}, we graph the top 10 highest performing templates, represented by the predicate, noun, and comparative in the sentence. We find that several templates do exhibit above chance performance, though accuracy is still generally low and confidence intervals are quite large (each template has dozens of examples). This finding indicates the models may have some limited semantic knowledge of \letalone, but do not encode an abstract \letalone construction. Instead, the construction is context dependent, and the intended meaning is only accessed alongside some specific lexical items and semantic frames.

\begin{figure*}
    \centering
    \includegraphics[width=\textwidth]{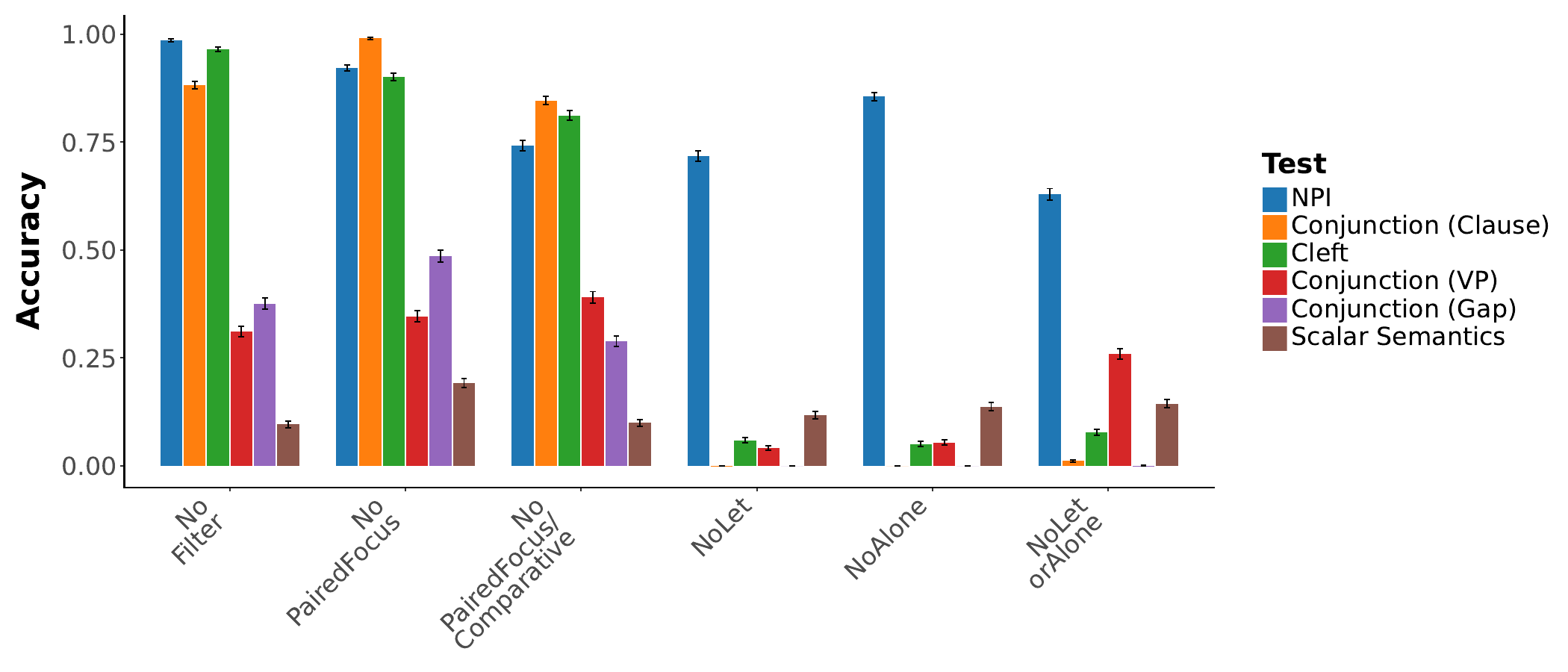}
    \caption{\textbf{Filtered Pretraining Results}. Accuracies are calculated according to Equation 3. Error bars are 95\% confidence intervals over the mean accuracies across two randomly seeded runs. \Cref{tab:filt_res} presents the same data.}
    \label{fig:filtered}
\end{figure*}

\subsection{Discussion}
 
These results, combined with those outlined in \cref{sec:form_results}, point to a strong divide between formal and functional competence regarding the \textsc{let-alone} construction for our human-scale models. This finding aligns well with past work on constructions \citep{Weissweiler_Hofmann_Köksal_Schütze_2022}, which has shown that syntactic competence often far outpaces semantic competence in controlled environments for a given construction.

Most constructionist accounts of language contend that constructions are stored as form\slash meaning pairings in the human mind, and posit that form and meaning are learned in conjunction by humans \citep{Goldberg_2006}. The apparent lack of functional knowledge of \letalone that we observe is compatible with a formal vs.~functional distinction of language competence in language models \citep{Mahowald_Ivanova_Blank_Kanwisher_Tenenbaum_Fedorenko_2024}.


\section{Experiment 3: Filtered Pretraining}\label{sec:exp3-filtered}

Having shown that language models have at least some sensitivity to the formal constraints, we now test how this capability is impacted by filtering the pretraining dataset to exclude \textsc{let-alone} and related constructions. We experiment with 5 ``filtered pretraining'' \citep{patilFilteredCorpusTraining2024} scenarios:

\noindent\textbf{Excluding all paired-focus constructions} (``let alone'', ``much less'', ``never mind'', ``not to mention''). These paired-focus constructions generally follow many of the same syntactic constraints as \textsc{let-alone}. These constructions combine to account for 2312 total sentences in the BabyLM train set.

\noindent\textbf{Excluding paired focus AND comparatives.} Since semantically, \textsc{let-alone} is thought to be related to comparative constructions, simple comparatives ``more than'' and ``less than'' are removed as well. This accounts for $\approx$70k sentences, or roughly 0.5\% of the entire BabyLM 100M train set. 

\noindent\textbf{Excluding all instances of \textit{let}.} This means \textit{let} is not seen as a token during pretraining. This accounts for $\approx$165k sentences in BabyLM train.\footnote{Inflectional variants like ``letting'' are not removed.}  

\noindent\textbf{Excluding all instances of \textit{alone}.} This excludes $\approx$16,000 sentences in BabyLM train.

\noindent\textbf{Excluding all instances of \textit{let} AND of \textit{alone}.} This means that neither token is seen (in any context) during pretraining ($\approx$180k sentences).

\noindent For all filtering, we use case-insensitive regular expression query matching over the BabyLM training corpus to remove examples. If a target construction is found, then the entire sentence containing that construction is removed from the training corpus. We use SLOR as the evaluation metric as in 
previous experiments, and calculate unigram probabilities independently for each filtered pretraining set. We test all filtered models on the minimal pair datasets from experiments in \cref{sec:exp1-formal} and \cref{sec:exp2-semantic}.

\begin{table*}[]
    \centering\smaller
    \begin{tabular}{ccccccc}
        \toprule
        \textbf{Filtering} & \textbf{NPI} & \textbf{Conjunction} & \textbf{Cleft} & \textbf{Conjunction} & \textbf{Conjunction} & \textbf{Scalar} \\
        \textbf{Scenario} & & \textbf{(Clause)} & & \textbf{(VP)} & \textbf{(Elided VP)} & \textbf{Semantics} \\
        \midrule
        NoFiltering & 98.6 $\pm$ 0.3\% & 88.1 $\pm$ 0.8\% & 96.5$\pm$ 0.5\% & 31.1 $\pm$ 1.3\% & 37.5 $\pm$ 1.3\% & \hphantom{0}4.9 $\pm$ 0.3\% \\
        NoPairedFocus & 91.9 $\pm$ 0.7\% & 97.6 $\pm$ 0.4\% & 93.7 $\pm$ 0.7\% & 39.8 $\pm$ 1.3\% & 58.1 $\pm$ 1.3\% & \hphantom{0}9.3 $\pm$ 0.4\% \\ 
        NoPairedFoc/Comp & 84.9 $\pm$ 0.9\% & 92.1 $\pm$ 0.7\% & 88.9 $\pm$ 0.8\% & 37.0 $\pm$ 1.3\% & 42.4 $\pm$ 1.3\% & 11.2 $\pm$ 0.5\% \\
        NoLet & 71.8 $\pm$ 1.2\% & \hphantom{0}0.0 $\pm$ 0.0\% & \hphantom{0}5.9 $\pm$ 0.6\% & \hphantom{0}4.1 $\pm$ 0.5\% & \hphantom{0}0.0 $\pm$ 0.0\% & \hphantom{0}1.8 $\pm$ 0.2\% \\
        NoAlone & 85.5 $\pm$ 0.9\% & \hphantom{0}0.0 $\pm$ 0.0\% & \hphantom{0}5.0 $\pm$ 0.5\% & \hphantom{0}5.4 $\pm$ 0.6\% & \hphantom{0}0.0 $\pm$ 0.0\% & \hphantom{0}3.8 $\pm$ 0.3\% \\
        NoLetorAlone & 62.9 $\pm$ 1.3\% & \hphantom{0}0.0 $\pm$ 0.0\% & \hphantom{0}7.7 $\pm$ 0.7\% & 25.9 $\pm$ 1.2\% & \hphantom{0}0.1 $\pm$ 0.0\% & \hphantom{0}6.4 $\pm$ 0.4\% \\

        \bottomrule
    \end{tabular}
    \caption{\textbf{Filtered Pretraining Results}. \Cref{fig:filtered} visualizes the same data.}
    \label{tab:filt_res}
\end{table*}

\subsection{Results}

\begin{figure*}
    \centering
    \includegraphics[width=.9\textwidth]{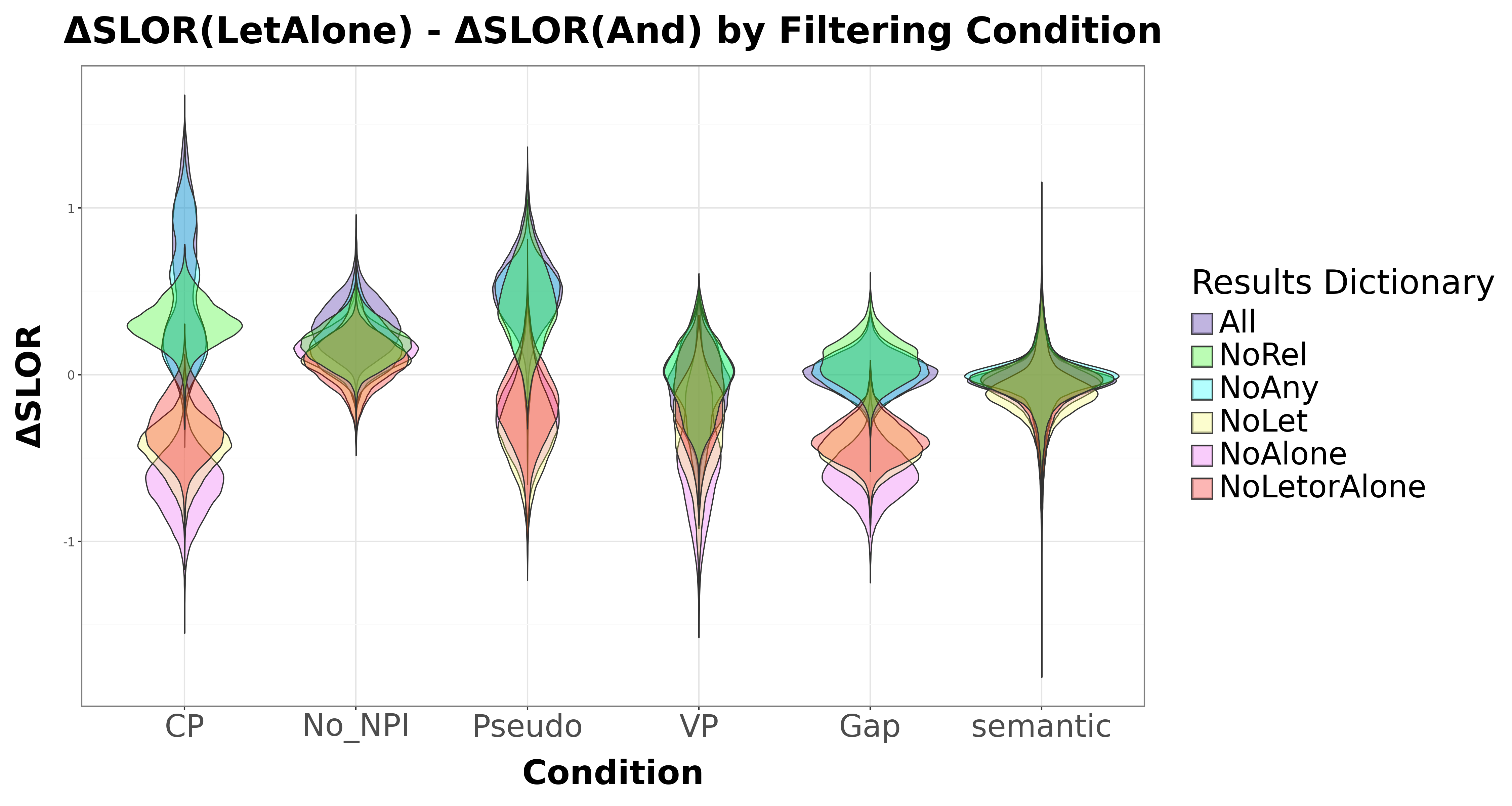}
    \caption{\textbf{$\Delta$SLOR(Let-Alone) $-$ $\Delta$SLOR(And) for Filtered Pretraining Conditions}. Positive Differences indicate sensitivity to the construction. Differences are generally still positive after removing direct evidence from related constructions (NoRel, NoAny) generally, but become negative when all `let' or `alone' tokens are removed.}
    \label{fig:filtered_viol}
\end{figure*}

We visualize the accuracies for each filtering scenario in \cref{fig:filtered}. We additionally visualize the changes in $\Delta$SLOR due to filtering in \cref{fig:filtered_viol}. Filtering out paired-focus constructions, including \letalone, seems to have little impact on the performance on formal tests, and we observe similarly high performance when additionally filtering out simple comparatives. However, when filtering out \textit{let} or \textit{alone}, performance drops substantially.

Interestingly, even when filtering out both \textit{let} and \textit{alone}, performance on the NPI constraint remains nontrivial. This likely points to some other heuristic that allows models to solve the task without any knowledge of the construction. We hypothesize that because negating a sentence with \textit{and} sometimes results in a sentence with \textit{or}, \textit{and} may be mildly biased against negative contexts.

\subsection{Discussion}

This work has shown that BabyLM scale models are sensitive to several formal constraints on \letalone. This sensitivity remains even when all instances of \letalone, related paired focus constructions, and even simple comparatives are removed from training, meaning the models are learning from indirect evidence of some kind beyond \letalone or seemingly related constructions. We hypothesize that the syntactic patterns tested here, while seemingly idiosyncratic to \letalone and similar constructions, are likely related to more general patterns which are better represented in pretraining data and thus facilitate learning \citep{Potts_2024}. In our case, the indirect evidence that models may be using is that of the manipulations that we test. For example, even without observing \letalone, our BabyLMs have observed cleft constructions, and may have learned that there is a restricted set of phrase types that can be clefted which does not include \letalone. Since all of our syntactic tests rely on combining \letalone with more common syntactic constructions, it is possible that robust representations of these interacting constructions allows for strong model performance on our tests.

We observe that performance degrades sharply when all \textit{let} or all \textit{alone} tokens are removed from pretraining. These results seem to indicate that LMs are using some compositionality between the embeddings of \textit{let} and \textit{alone} to arrive at the meaning of the construction overall. This is somewhat counterintuitive in that Construction Grammar theory would not necessarily predict such a strong link between lexical items and a construction in which they are used far outside of their canonical distributions. However, we also note that SLOR as a metric inherently involves normalizing language model scores by the probability of a string from a unigram language model. Thus, removing \textit{let} and \textit{alone} entirely has a substantial effect on the unigram-based probability, and ultimately may explain the large drop in performance when \textit{let} and \textit{alone} are filtered. We leave open the possibility of replacing SLOR with a more complex function \citep[e.g.~MORCELA;][]{tjuatja-etal-2025-goes} for future work.

\section{Related Work}\label{sec:related}

\paragraph{Human-Scale LMs.}
There is increasing interest in designing smaller scale LMs which could potentially be more informative to human congition \citep{dupouxCognitiveScienceEra2018}. 
The BabyLM challenge \citep{warstadtFindingsBabyLMChallenge2023} was created to address this interest and has resulted in robust human-scale models \citep{georgesgabrielcharpentierNotAllLayers2023}. Furthermore, growing evaluation of constructional knowledge in BabyLMs has yielded promising results \citep{misraLanguageModelsLearn2024,bunzeck-etal-2025-construction,rozner2025babylmsconstructionscausalinterventions}. Beyond BabyLM, researchers have endeavored to create smaller scale LMs using a variety of training corpora, including pretraining on the British National Corpus \citep{bnc2007british} and achieving respectable performance on a variety of syntactic and understanding benchmarks \citep{Samuel_Kutuzov_Øvrelid_Velldal_2023}.

\paragraph{Constructions in LMs.}
This present work follows in a line of research which seeks to test language model understanding of ``constructions'' as defined by Construction Grammar \citep{Goldberg_1995,Croft_2001}, of which \letalone is just one construction out of many. Starting with CxGBERT \citep{Tayyar_Madabushi_Romain_Divjak_Milin_2020}, there have been a flurry of papers showing that LMs learn a variety of constructions, including abstract argument structures \citep{Li_Zhu_Thomas_Rudzicz_Xu_2022}. \citet{Weissweiler_Hofmann_Köksal_Schütze_2022} is of particular relevance to this work, as they perform a paired syntactic/semantic analysis of a rare construction, though not at human-scale. Using probing tasks, they show that BERT-scale LMs recognize the syntax of the \textsc{comparative-correlative} but not its semantics. Furthermore, there have been several works that have shown LM reasoning and semantic capabilities are limited when confronted with rare constructions \citep{zhou-etal-2024-constructions,bonialConstructionGrammarCorpus2024b}, though results from other studies are more promising \citep{Mahowald_2023,Potts_2024,Rozner_Weissweiler_Mahowald_Shain_2025,scivetti-schneider-2025-construction}.

This present work most directly builds off of \citet{misraLanguageModelsLearn2024}, who show that BabyLM scale models are sensitive to the formal properties of the AANN construction. They also are the first to apply the ``filtered pretraining'' \mbox{\citep{patilFilteredCorpusTraining2024}} technique to construction-based investigations, and show that language models are able to remain sensitive to properties of AANNs even when they don't observe them during training. The findings of this present work on \textsc{let-alone} largely support this claim that properties of constructions can be learned without direct observation in training, which has been shown to be true (to some extent) even for more frequent constructions like the dative alternation \citep{Yao_Misra_Weissweiler_Mahowald_2025}.

\paragraph{Let Alone.}
\citet{bonialConstructionGrammarCorpus2024b}, \citet{Scivetti_Torgbi_Blodgett_Shichman_Hudson_Bonial_Madabushi_2025}, and \citet{Rozner_Weissweiler_Mahowald_Shain_2025} are the only past works (to our knowledge) which have specifically targeted language model understanding of \letalone in some capacity. \citet{Rozner_Weissweiler_Mahowald_Shain_2025} introduce \textit{global affinity} and \textit{local affinity} as metrics to quantify the extent to which constructional information is approximated by an LM's output distribution. Using RoBERTa, they show that for corpus instances of \letalone, both \textit{let} and \textit{alone} have high global affinity scores, indicating that the model has at least some distributional knowledge that links the two words as part of a construction. Focusing on natural corpus data from the CoGS dataset, \citet{bonialConstructionGrammarCorpus2024b} find that \letalone can be distinguished from distractor constructions at a much higher accuracy by LLMs when compared to fully abstract constructions. \citet{Scivetti_Torgbi_Blodgett_Shichman_Hudson_Bonial_Madabushi_2025} extend this work 
by refashioning the corpus examples into a natural language inference (NLI) dataset, finding that LLMs can perform NLI with very high accuracy for examples which target the semantics of \letalone. Our work diverges from the prior work on \letalone in that we 1)~focus on a templatically generated dataset as opposed to corpus data, allowing us to test much finer-grained properties of the construction, and 2)~use minimal pair--based evaluations to test both \textit{form} and \textit{meaning} on human-scale models.

\section{Conclusion}
This work has shown that BabyLM models can learn various formal properties of \letalone. This formal competence is maintained even in the absence of direct evidence of \letalone and related constructions. Such a result points to the crucial nature of indirect evidence for learning a construction as rare as \letalone. On the other hand, BabyLM models seem to have very little grasp of \letalone's meaning. This result underscores the importance of considering both formal and functional competence when assessing LM capabilities regarding a specific linguistic phenomenon, and doing so in a controlled manner. While the formal capabilities of BabyLM models are promising, insofar as we consider meaning to be a central part of language,
our results cast doubt on the proposition that robust semantics---including of rare constructions---can be learned from form alone via human-scale pretraining.

\section*{Limitations}

In this work, we focus on a single construction, \letalone. 
Future work is needed to determine the extent to which models can learn the form and meaning of various constructions from human-scale pretraining data. Secondly, we only test one type of model architecture. This architecture has been found to be robust at learning from human-scale data in past work, but we cannot be sure that the results would hold for other architectures. Finally, this work only analyzes a single construction in a single language, English, while more work is needed to test a wider variety of rare constructions across languages. 

\section*{Ethics and Risks}
We do not foresee any major ethical concerns or risks associated with this work. All evaluation data is templatically created and thus free of any sensitive or offensive content. The dataset released here is designed for research into the \letalone construction, and the authors do not foresee that it could be deployed for any malicious purpose.

\section*{Acknowledgments}

We thank the anonymous reviewers, ACs, and SACs for EMNLP and the ARR May Cycle. We also thank members of the NERT and PiCoL labs for their useful comments on early versions of this work. This research was supported in part by NSF award IIS-2144881.

\bibliography{mypaper}

\appendix

\section{Pretraining Hyperparameters}\label{hyper}

\cref{tab:model_architecture} reports the hyperparameters for BabyLM pretraining, which were kept consistent across all runs. This is the same as those reported in \citet{misraLanguageModelsLearn2024}, except that they tune the learning rate between the options of 1e-4, 3e-4, 1e-3, and 3e-3. We opt for 1e-4 after preliminary runs showed that higher learning rates led to development loss bottoming out much before the end of 20 epochs of training. In total, 12 complete pretraining runs were performed, totaling roughly 100 hours runtime on a  NVIDIA A100 GPU. 

\begin{table}[h]
    \centering
    \begin{tabular}{l c}
        \toprule
        Model Architecture & OPT \citep{Zhang_Roller_Goyal_Artetxe_Chen_Chen_Dewan_Diab_Li_Lin_etal._2022} \\
        Embed Size & 768 \\
        FFN Dimension & 3,072 \\
        Num. Layers & 12 \\
        Attention Heads & 12 \\
        Vocab Size & 16,384 \\
        Max. Seq. Length & 256 \\
        Batch Size & 32 \\
        Warmup Steps & 32,000 \\
        Epochs & 20 \\
        Learning Rate & 1e-4 \\
        Total Parameters & 97M \\
        Training Size & 100M tokens \\
        \bottomrule
    \end{tabular}
    \caption{Model hyperparameters for all OPT Models}
    \label{tab:model_architecture}
\end{table}

\section{GPT Prompting Experiment}\label{gpt_exp}
For the semantic experiment (see \cref{sec:exp2-semantic}), we reformulated our task as a prompt-based experiment for input into GPT-4.1. Since this result serves primarily as a comparison point to BabyLM models, we do not perform prompt optimization. Thus, the reported GPT-4.1 accuracy should not be taken as an upper limit on LLM performance on this dataset. GPT-4.1 was accessed through the OpenAI API. The cost of this experiment was roughly \$8 USD. \cref{tab:gpt_prompt} reports the prompt used for GPT-4.1 experiments.

\begin{table}[t]
  \centering
  \begin{tabular}{@{}p{0.97\linewidth}@{}}
    \toprule
    \textbf{Prompt template fed to GPT‑4.1}\\
    \midrule
    \begin{minipage}[t]{\linewidth}\small\ttfamily
You are a linguistic annotator who must understand the `let alone`
construction. Please read the following text, and consider its meaning.
Then choose the sentence that is most likely to be true, given the text.

Text:
\texttt{\{text\}}

Question:
Which of the following sentences is most likely to be true, given the above
text?

Choices:
\texttt{\{formatted\_choices\}}

Please respond with only the letter of the correct choice (e.g.\ A or B).
    \end{minipage}\\
    \bottomrule
  \end{tabular}
  \caption{Exact prompt template used in our experiments. Curly‑braced items
           (\texttt{\{text\}}, \texttt{\{formatted\_choices\}}) are filled in
           programmatically with items from our semantic dataset.}
  \label{tab:gpt_prompt}
\end{table}

\end{document}